\documentclass[letterpaper]{article} % DO NOT CHANGE THIS
\usepackage{aaai24}  % DO NOT CHANGE THIS
\usepackage{times}  % DO NOT CHANGE THIS
\usepackage{helvet}  % DO NOT CHANGE THIS
\usepackage{courier}  % DO NOT CHANGE THIS
\usepackage[hyphens]{url}  % DO NOT CHANGE THIS
\usepackage{graphicx} % DO NOT CHANGE THIS
\usepackage{natbib}  % DO NOT CHANGE THIS AND DO NOT ADD ANY OPTIONS TO IT
\usepackage{caption} % DO NOT CHANGE THIS AND DO NOT ADD ANY OPTIONS TO IT
\usepackage{algorithm}
\usepackage{algorithmic}

\usepackage{amsmath}
\usepackage{amssymb}
\usepackage{cite}
\usepackage{amsmath}
\usepackage{mathrsfs}
\usepackage{bm}
\usepackage{multirow}
\usepackage{bbding}

\usepackage{subfigure}
\usepackage{color}
\usepackage{makecell}
\usepackage{newfloat}
\usepackage{listings}
\DeclareCaptionStyle{ruled}{labelfont=normalfont,labelsep=colon,strut=off} % DO NOT CHANGE THIS
\floatstyle{ruled}
\newfloat{listing}{tb}{lst}{}
\floatname{listing}{Listing}
\title{Boosting Few-Shot Learning via Attentive Feature Regularization}
\title{Boosting Few-Shot Learning via Attentive Feature Regularization}
\author{
    Xingyu Zhu\textsuperscript{\rm 1, 2}, 
    Shuo Wang\textsuperscript{\rm 1, 2}\thanks{Corresponding author}, 
    Jinda Lu\textsuperscript{\rm 1, 2}, 
    Yanbin Hao\textsuperscript{\rm 1, 2}, 
    Haifeng Liu\textsuperscript{\rm 3}, 
    Xiangnan He\textsuperscript{\rm 1, 2}
}
\affiliations{
    \textsuperscript{\rm 1}Department of Electronic Engineering and Information Science, University of Science and Technology of China;\\
    \textsuperscript{\rm 2}MoE Key Laboratory of Brain-inspired Intelligent Perception and Cognition, University of Science and Technology of China;\\
    \textsuperscript{\rm 3}Brain-Inspired Technology Co., Ltd.
    \{xingyuzhu, lujd\}@mail.ustc.edu.cn, \{shuowang.edu, xiangnanhe\}@gmail.com, \\
    haoyanbin@hotmail.com, haifeng@leinao.ai
}

\usepackage{bibentry}
\begin{document}

\maketitle

\begin{abstract}
Few-shot learning (FSL) based on manifold regularization aims to improve the recognition capacity of novel objects with limited training samples by mixing two samples from different categories with a blending factor. However, this mixing operation weakens the feature representation 
due to the linear interpolation and the overlooking of the importance of specific channels.
To solve these issues, this paper proposes attentive feature regularization (AFR) which aims to improve the feature representativeness and discriminability. In our approach, we first calculate the relations between different categories of semantic labels to pick out the related features used for regularization. Then, we design two attention-based calculations at both the instance and channel levels. These calculations enable the regularization procedure to focus on two crucial aspects: the feature complementarity through adaptive interpolation in related categories
and the emphasis on specific feature channels.
Finally, we combine these regularization strategies to significantly improve the classifier performance. Empirical studies on several popular FSL benchmarks demonstrate the effectiveness of AFR, which improves the recognition accuracy of novel categories without the need to retrain any feature extractor, especially in the 1-shot setting. Furthermore, the proposed AFR can seamlessly integrate into other FSL methods to improve classification performance.
% (63.17\% to 69.78\% on the Mini-ImageNet dataset). 
\end{abstract}

\section{Introduction}
In recent years, convolutional neural networks (CNNs) have demonstrated remarkable capabilities on various visual classification tasks, particularly provided with sufficient training data. However, collecting and labeling such datasets is a time-consuming and expensive procedure. As a remedy to address this challenge, few-shot learning (FSL) is proposed to classify a novel object with a scarcity of labeled data. \cite{ye2020few,peng2019few,wang2020large}.

\begin{figure}[!t]
\centering
\includegraphics[width=0.45\textwidth]{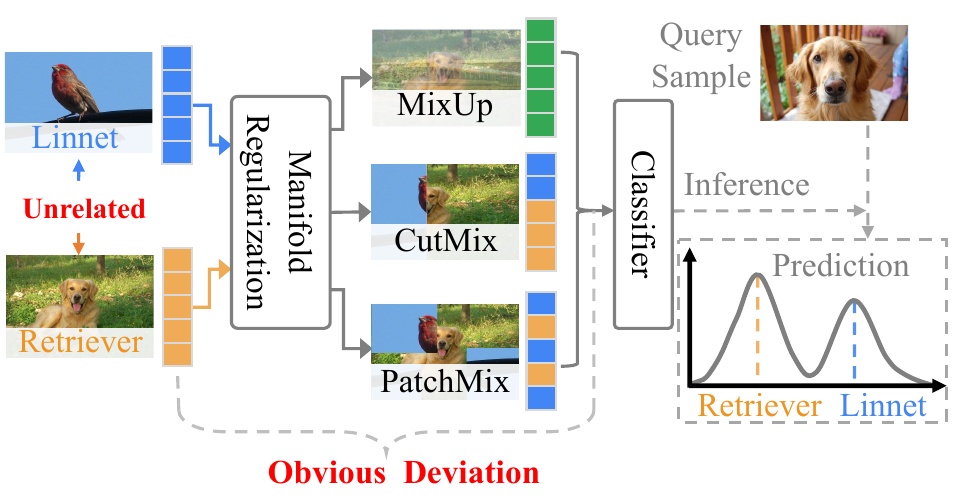}
\caption{The analysis of manifold regularization methods.}
\label{fig:fig1}
\vspace{-0.2cm}
\end{figure}

The conventional solution of FSL involves using a CNN trained on the base categories to directly extract the global features of novel objects \cite{hariharan2017low,wang2018low}. It aims to yield a transferable feature representation (textures and structures) to describe a novel category. Subsequently, these features are employed to train a classifier for recognizing novel objects. Manifold
 regularization \cite{rodriguez2020embedding, deutsch2017zero, velazquez2022closer} is a popular strategy to improve classification performance. These methods involve mixing two samples (features) and their labels from randomly selected categories to 
 generate a regularized feature. However, the mixing operation with randomness is easy to weaken the representation ability \cite{GuoMZ19, Remix}. This is primarily due to the direct interpolation without considering the complementarity of two features and the neglect of specific feature channels \cite{HouLW17, ShiWW23a, luo2022channel, zhu2023not}, which in turn impacts the distribution of prediction results.
 As illustrated in Figure \ref{fig:fig1}, given a novel sample of ``Retriever'' and another randomly picked out sample ``Linnet'', the manifold regularization methods, \textit{e.g.}, Mixup \cite{zhang2018mixup}, CutMix \cite{yun2019cutmix}, and PatchMix \cite{liu2021learning}, interpolate their images and labels to train the classifier for predicting both categories. It's evident that the ``Retriever'' and the ``Linnet'' are unrelated in terms of both vision and semantics. Consequently, the regularized features deviate from the novel feature ``Retriever" (as indicated by the five yellow squares in the lower-left corner of Figure 1). This deviation leads to an increase in the prediction score for ``Linnet" and results in misclassification. This deviation leads to an increase in the prediction score of the ``Linnet'' and limits the classification results.

To address the aforementioned issue arising from manifold regularization, we first incorporate semantics to select categories related to the novel categories from the base set. This idea aligns with the prior work \cite{wang2020large, peng2019few, wang2022multi}, where semantic knowledge not only strengthens visual features but also aids help classifier in capturing discriminative patterns. However, it's worth noting that such methodologies necessitate greater prior semantic information during the training process, which leads to increased model size and longer training times.
% Previous work \cite{wang2020large, peng2019few, wang2022multi} also utilizes the semantic knowledge, which leverages the textual label not only to strengthen the visual feature but also assists the classifier in capturing the discriminative pattern. However, it's worth noting that such methodologies necessitate more prior semantics involved in the training process, which increases the model size and training time.
Different from the previous approaches, our method solely relies on semantic labels to select relevant base categories during the data preprocessing stage. This purposeful selection, in contrast to the random selection in manifold regularization, enables the classifier to better concentrate more effectively on the novel content during the training stage. Besides, we also exploit the feature complementarity from similar categories and the discriminability of specific feature channels, which can both provide distinctive patterns for classification \cite{LiuZZCZ19, ShiWW23a}. Building on the above analysis, we propose two attention-based calculations at the instance and channel levels, respectively.
% ，which provide distinct information to each other \cite{LiuZZCZ19, ShiWW23a}, and the discriminability of specific feature channels, which affords the distinctive patters to the categories. Building on the above analysis, we propose two attention-based calculations at the instance and channel levels.
% only utilize the semantic labels to select the related base categories in the data preprocessing stage. This purposeful selection contrasts with random selection in manifold regularization helps the classifier be better able to focus on
% the novel content during the training. 
% Then, we exploit the collaboration between the related categories and the discriminability of the specific feature channels, which can both provide distinctive patterns for the categories \cite{HouLW17, luo2022channel, ShiWW23a}. Based on the above analysis, we propose two attention-based calculations at the instance and channel levels.
% % Furthermore, to exploit the collaboration between related categories and the discriminability of the specific feature channels, we propose two attention-based calculations at the instance and channel levels.

The instance attention is designed to adaptively leverage the collaborative components of the 
selected base categories guided by their relevance to heighten the novel feature representativeness.
Specifically, we first analyze the semantic similarity of the selected base categories related to the novel category, then calculate the attention scores between the selected base samples and the given novel sample to measure their importance. Finally, these attention scores are then employed in the reweighting of selected samples. Instance attention exploits the collaboration between the related categories through adaptive interpolation, which avoids the irrelevant components of the base categories and consequently improves the representation of the novel samples.

For channel calculations, we aim to emphasize the specific feature channels that signify the discriminative patterns.
Specifically, we calculate the scores as channel importance weights from the regularized features output from the instance attention. These weights are then applied to each channel of features in regularization, aiding the classifier in identifying the representative content of the novel samples. This channel attention mechanism allows for more efficient and focused exploration of novel category information within the feature channels, which enhances the discriminability of the final feature representation.

The proposed procedures, defined as Attentive Feature Regularization (AFR), all operate on features and can be easily applied to existing pre-trained feature extractors.
The main contributions of our method are as follows.
\begin{enumerate}
\item We propose instance-level attention with semantic selection to improve the feature representativeness, which 
leverages the complementarity of the related base categories to enhance the novel categories.
\item We design channel-level attention to enhance the feature discriminability by measuring the importance of different channels, which helps the classifier focus on the representative content of the novel sample.
\item Our method achieves state-of-the-art performance on three popular FSL datasets and can also be used to improve the performance of the classifier in other FSL methods without training feature extractors.
\end{enumerate}

\begin{figure*}[t]
\centering
\includegraphics[width=0.98\linewidth]{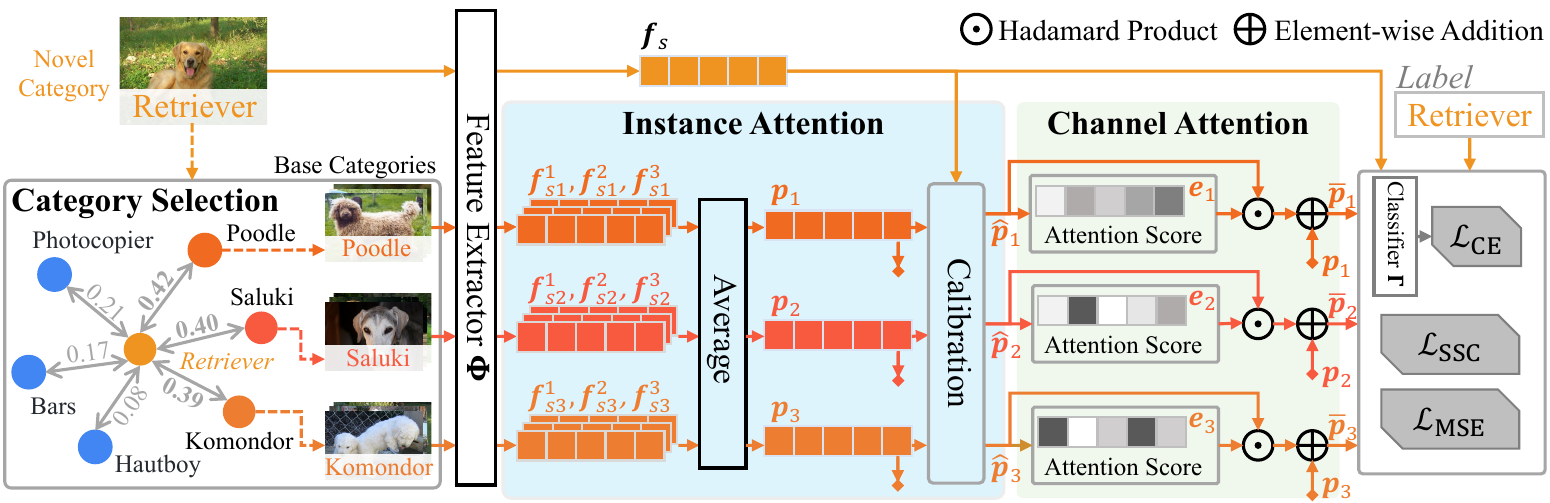}
\caption{The overview of attentive feature regularization (AFR), where $\mathcal{L}_{\rm CE}$, $\mathcal{L}_{\rm SC}$, and $\mathcal{L}_{\rm MSE}$ are three losses.}
\label{fig:overview}
\end{figure*}

\section{Related Work}
In this section, we first briefly introduce common solutions for FSL tasks and corresponding regularization strategies. Subsequently, we list the applications of recent attention-based methods. Finally, we enumerate the differences between our methods and those of related methods.
% \subsection{Traditional Few-Shot Learning}
% Recent advances in Few-Shot Learning (FSL) have shown promising performance by developing the robust feature extractor in a good structural design or model training \cite{peng2019few, liu2021learning}.
% These methods involve learning how to learn from
% multiple FSL tasks with a set of base categories to train a meta-learner capable of adaption to novel categories, and they can be broadly categorized into two types: optimization-based method and metric-based method. For optimization-based methods, such as Model-Agnostic Meta-Learning (MAML) \cite{finn2017model}, Reptile\cite{nichol2018reptile}, and so on. They focus on learning an initial set of model parameters that can be easily fine-tuned to adapt to new tasks. For metric-based methods, including ProtoNet\cite{snell2017prototypical}, MatchNet\cite{vinyals2016matching}, aims to train a neural network that can effectively bring samples from the same category closer to each other in the feature space, while pushing samples from different categories farther apart.

\subsection{Knowledge Transfer in Few-Shot Learning}
Recent advances in Few-Shot Learning (FSL) have demonstrated promising performance by transferring the knowledge from the base categories to the novel categories \cite{Limargin, li2019large, wang2022multi, Lumm}. These methods leverage semantic knowledge to provide additional information for refining visual features or enriching the supervision during classifier training. For example, the method in \cite{li2019large} clusters hierarchical textual labels from both the base and novel categories to improve the feature extractor training. Wang \textit{et al.} proposed a multi-directional knowledge transfer (MDKT) method which integrates the visual and textual features through a bidirectional knowledge connection. The work described in \cite{Lumm} employs the semantics to explore the correlation of categories to hallucinate the additional training samples.

\subsection{Regularization in Few-Shot Learning	}
Recently, manifold regularization\cite{devries2017improved, zhang2018mixup, verma2019manifold, yun2019cutmix, liu2021learning} has been used in FSL tasks, 
which is simply based on mixture and mask operation and can improve the classification performance. 
The simplest method is CutOut \cite{devries2017improved}, which randomly masks out square regions of input during training and improves the performance of the networks. Based on CutOut, many other manifold regularization methods have been developed,
\textit{i.e.}, MixUp \cite{zhang2018mixup,verma2019manifold}, CutMix \cite{yun2019cutmix}, PatchMix \cite{liu2021learning}. Specifically, MixUp mixes two samples by interpolating both the image and the labels. In CutMix, patches are cut and pasted among training features, where ground truth labels are also mixed proportionally to the area of the patches. PatchMix is similar to CutMix and uses mixed images for contrastive learning. 

\subsection{Attention in Few-Shot Learning}
In the field of FSL, attention mechanisms \cite{VaswaniSPUJGKP17} have been widely widespread due to their ability to highlight the important parts of inputs by measuring similarities. This enables the network to focus on critical content for specific tasks \cite{hou2019cross, kang2021relational, chikontwe2022cad}. For instance, Hou \textit{et al.} proposed a cross-attention (CAM) method to model the semantic relevance between class and query features, leading to adaptive localization of relevant regions and generation of more discriminative features \cite{hou2019cross}.
The work in \cite{kang2021relational} computes the cross-correlation between two representations and learns to produce co-attention between them. It improves the classification accuracy by learning cross-correlational patterns and adapting ``where to attend'' concerning the images given in the testing stage. Moreover, the method \cite{ye2020few} integrates the entire Transformer module including attention mechanisms (FEAT) to adapt the features for FSL tasks. The authors in \cite{lai2022tsf}
propose the method named transformer-based Semantic Filter (tSF) which defines the additional learnable parameters to filter the useful knowledge of the whole base set for the novel category.
The recently proposed CAD \cite{chikontwe2022cad} employs self-attention operations to cross-attend support and query embeddings, effectively reweighting each instance relative to the others.

Based on the analysis of the related work, our method belongs to the manifold regularization methods.
The methods most related to ours are the recently proposed Manifold Mixup in \cite{verma2019manifold} and CAM in \cite{hou2019cross}. Our method differs from theirs in two aspects. First, we introduce semantic knowledge to purposefully select samples for regularization and keep the label of the regularized feature the same as the novel feature to avoid introducing other unrelated supervisions for training. Second, we design two attention calculations to enhance collaboration and improve the discriminability of features, which helps the classifier focus on the distribution of novel categories rather than associate the support and query samples during the testing stage \cite{hou2019cross}. Besides, our approach directly applies to the features and has a lower computational complexity.

\section{Method}
In this section, we elaborate on our attentive feature regularization (AFR). First, we briefly revisit the preliminaries of the FSL tasks and an overview of our framework. Second, we delve into the details of our semantic selection process and different attention calculations. Finally, we describe the training and inference procedures of our approach.

\subsection{Preliminaries}
The data for the few-shot learning task is divided into three parts: base set $\mathcal{D}_{\rm base}$, support set $\mathcal{D}_{\rm support}$, and query set $\mathcal{D}_{\rm query}$. The base set $\mathcal{D}_{\rm base}$ has large-scale labeled samples (\textit{e.g.}, about hundreds of samples in one category) used for training the feature extractor. The categories of these samples are denoted as $\mathcal{C}_{\rm base}$ and provide valuable prior knowledge as known contents to describe other samples. The support set $\mathcal{D}_{\rm support}$ and the query set $\mathcal{D}_{\rm query}$ share the same set of categories, called $\mathcal{C}_{\rm novel}$, which is disjoint with that of the base set $\mathcal{C}_{\rm base}$. 
The goal of few-shot learning is to construct a classifier
 using training samples from both the base set and the support set, capable of accurately classifying the samples in the query set.
For the training samples from $\mathcal{D}_{\rm support}$, there are total $N$ categories that are randomly sampled from $\mathcal{C}_{\rm novel}$, and each category provides $K$ samples. This process is known as the $N$-way-$K$-shot recognition problem.

The overview of our framework is depicted in Figure \ref{fig:overview}. First, we use the semantic knowledge to select the related base categories to a given novel sample and extract features of all these samples by a pre-trained CNN. Second, we design instance attention and channel attention to regularize these features. Third, we design three losses to constrain the regularization procedure and train a classifier.

\subsection{Attentive Feature Regularization}
Textual knowledge uses semantic description to express each category. It provides the direct relations between the categories. To avoid bringing irrelevant noise to influence the classifier training, we directly calculate the relations between these descriptions before regularization. Specifically, we first use the word2vec embedding method \cite{li2019large} to express these descriptions into the feature. Then, given feature of a support category as $\bm{t}_s$, we calculate the relations $\mathcal{R}^s=\{r^s_i\}_{i=1}^{|\mathcal{C}_{\rm base}|}$ between $\bm{t}_s$ and the other descriptions $\{\bm{t}_i\}_{i \in \mathcal{C}_{\rm base}}$ from base categories by similarity calculation: 
\begin{eqnarray}
r^s_i = \frac{\langle \bm{t}_s, {\bm{t}}_i \rangle}{\|\bm{t}_s\|_{2} \cdot \|\bm{t}_i\|_{2}},
\label{eq:selection}
\end{eqnarray}
where $\langle$$\cdot$, $\cdot$$\rangle$ is the inner product between two vectors. 

After obtaining the relation scores $\mathcal{R}^s$, we sort them and select the samples from the top-$\beta_s$ related categories denoted as $\mathcal{C}_{\rm \beta_s}$ for regularization. These semantically relevant features can provide a more relevant content supplement to the training and avoid bringing much irrelevant noise.

 Our regularization operates on the feature level. Therefore, we first represent the sample $\bm{I}$ into feature $\bm{f} = \bm{\Phi}(\bm{I})\in\mathbb{R}^{d}$ by extracting the output from the pre-trained visual model $\bm{\Phi}$ before the last prediction layer. The model $\bm{\Phi}$ is already trained on known images from the base set $\mathcal{D}{\rm base}$, and $d$ is the dimension of the feature. Then, we illustrate the different attention calculations used in our approach.

\begin{figure}[t]
\centering
\includegraphics[width=0.45\textwidth]{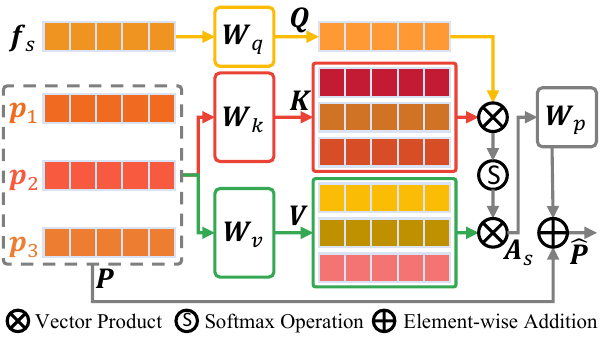}
\caption{The calculation of calibration in instance attention.}
\label{fig:instance_att}
\end{figure}

\subsubsection{Instance Attention}
Given a support feature $\bm{f}_s$ with its textual description feature $\bm{t}_s$, we first selected ${\rm \beta_s}$ categories from base categories as $\mathcal{C}_{\rm \beta_s}$ using Eq \eqref{eq:selection}. Then we compute the prototype of each selected category $\mathcal{C}_{\rm \beta_s}^i$ by averaging all the features corresponding to that category:
\begin{eqnarray}
% \bm{p}_i = \frac{1}{|\mathcal{C}_{\rm \beta_s}^i|}\sum\nolimits_{\bm{f}_j \in \mathcal{C}_{\rm \beta_s}^i} \bm{f}_j.\\
\bm{p}_i = \frac{1}{|\mathcal{C}_{\rm \beta_s}^i|}\sum\nolimits \bm{f}_j, \ j \in \mathcal{C}_{\rm \beta_s}^i
\end{eqnarray}
Therefore, the prototypes of whole $\mathcal{C}_{\rm \beta_s}$ categories can be constructed as $\bm{P} = [\bm{p}_1, \bm{p}_2, ..., \bm{p}_{|\mathcal{C}_{\rm \beta_s}|}]$, where $\bm{P} \in \mathbb{R}^{{|\mathcal{C}_{\rm \beta_s}|} \times d}$. We then design a calibration based on self-attention \cite{vaswani2017attention} to find relevant categories that can help describe the novel category. The details are shown in Figure \ref{fig:instance_att}. Specifically, we first design three attention matrixes $\bm{Q}$, $\bm{K}$, and $\bm{V}$ to capture the content similarities from novel features and prototypes of related categories:
\begin{eqnarray}
\bm{Q}= \bm{f}_s * \bm{W}_q, 
\bm{K}= \bm{P} * \bm{W}_k,
\bm{V}= \bm{P} * \bm{W}_v,
\end{eqnarray}
where $\bm{W}_q \in \mathbb{R}^{d\times d}$, $\bm{W}_k\in \mathbb{R}^{d\times d}$, $\bm{W}_v\in \mathbb{R}^{d\times d}$ are weights of calibration calculation. We then use these similarities to calibrate the prototypes of the related categories since the distribution of the base categories and the novel category belong to different spaces, where the amplitude $\bm{A}_s \in \mathbb{R}^{d}$ of calibration is calculated as:
\begin{eqnarray}
\bm{A}_s= {\rm softmax}(\frac{\bm{Q}*{\bm{K}}^\intercal}{\sqrt{d}}){\bm{V}}.
\end{eqnarray}
It measures the relations between the novel category and its related base categories from the feature space.

Thus, calibrated prototypes $\hat{\bm{P}}$ can be defined as:
\begin{align}
\begin{aligned}
\hat{\bm{P}} &= [\hat{\bm{p}}_1, \hat{\bm{p}}_2, ..., \hat{\bm{p}}_{|\mathcal{C}_{\rm \beta_s}|}], \\
&= \delta(\bm{A}_s * \bm{W}_p) + [\bm{p}_1, \bm{p}_2, ..., \bm{p}_{|\mathcal{C}_{\rm \beta_s}|}],
\end{aligned}
\label{eq:insatt}
\end{align}
where $\bm{W}_p \in \mathbb{R}^{d\times d}$ is weight matrix for the calibration calculation, and $\delta$ is ReLU function. The calibrated prototypes can better simulate the distribution of the novel category and improve the accuracy of feature regularization.

\subsubsection{Channel Attention}
Channels of features have different influences on classifiers \cite{yue2020interventional}. To identify the important content of the channels, we design a channel attention module inspired by SE-Net \cite{hu2018squeeze}. SE-Net terms the ``Squeeze-and-Excitation (SE)'' block to adaptively re-calibrate channel-wise feature responses by explicitly modeling interdependencies between channels in backbone training\cite{hu2018squeeze}. Thus, we introduce a similar operation into the feature analysis. Specifically, we design two fully connected (FC) layers to ``Squeeze-and-Excitation'' calibrated prototypes $\hat{\bm{P}}$:
\begin{eqnarray}
\bm{E_s} = \sigma({\bm{FC}}_2(\delta({\bm{FC}}_1(\hat{\bm{P}})))),
\end{eqnarray}
where $\sigma$ is the Sigmoid function, and we intentionally set the embedding size of ${\bm{FC}}_1$ to be smaller than that of ${\bm{FC}}_2$. This design enhances important content and weakens unrelated content in the features by controlling the size of ${\bm{FC}}_1$. To further fuse the channel attention with the prototypes, we set the size of $\bm{E_s}$ to be the same as $\hat{\bm{P}}\in\mathbb{R}^{{|\mathcal{C}_{\rm \beta_s}|} \times d}$ by controlling ${\bm{FC}}_2$ accordingly. In our fusion stage, we employ a residual structure to prevent vanishing gradients while improving the accuracy of the prototype representation:
\begin{eqnarray}
\bar{\bm{P}} = \bm{E_s} \odot \hat{\bm{P}} + {\bm{P}},
\label{equ_chanatt}
\end{eqnarray}
where $\odot$ is the Hadamard product. $\bar{\bm{P}}\in\mathbb{R}^{{|\mathcal{C}_{\rm \beta_s}|} \times d}$ not only close to the distribution of the novel category by calibrating but also captures the content related to the novel category by using channel attention. Therefore, we sample the features of $\bar{\bm{P}}$ as representations of the given novel category to enrich the training set in our few-shot learning task.

\subsection{Training and Inference}
Denoted the novel samples and their labels in a $N$-way-$K$-shot task as $\{\{\bm{f}_s^i, \bm{l}_s^i\}_{s=1}^N\}_{i=1}^K$, and the fused prototypes as $\{\bar{\bm{P}}_s = \{\bar{\bm{p}}_s^j\}_{j=1}^{\beta_s}\}_{s=1}^N$, we combine the given features and prototypes into one set $\{\{\bm{H}_s = \{{\bm{h}}_s^j\}\}_{s=1}^N\}_{j=1}^{K+\beta_s}$ to simplify the expressions in subsequent calculations, where ${\bm{H}_s} = [\bm{f}_s^1, \bm{f}_s^2, ...,\bm{f}_s^K, \bar{\bm{p}}_s^1, \bar{\bm{p}}_s^2, ..., \bar{\bm{p}}_s^{\beta_s}]$. We then design two
losses to constrain the distribution of regularized prototypes
and use cross-entropy (CE) loss to train the classifier.

First, we adopt the principles of self-supervised contrastive learning \cite{khosla2020supervised}, which aim to bring features of the same category closer together while pulling features of different categories apart. Thus, the supervised contrastive (SC) loss can be calculated as follows:
% \begin{small}
\begin{eqnarray}
% \mathcal{L}_{\rm SC} = \frac{1}{N}\frac{1}{|\bm{H}_s|}\sum_{s=1}^N\sum_{\substack{i,j=1\\i\neq j}}^{|\bm{H}_s|}\log\frac{\exp(\langle \bm{h}_s^i, \bm{h}_s^j\rangle /\tau)}{\sum\limits_{\forall\bm{h}_p\notin\bm{H}_s}\exp(\langle \bm{h}_s^i, \bm{h}_p\rangle/\tau)},
\mathcal{L}_{\rm SC} \!=\! \frac{1}{{N}|\bm{H}_s|}\sum_{s=1}^N\sum_{\substack{i,j=1\\i\neq j}}^{|\bm{H}_s|}\log\frac{\exp(\langle \bm{h}_s^i, \bm{h}_s^j\rangle /\tau)}{\sum\limits_{\forall\bm{h}_p\!\notin\bm{H}_s}\!\exp(\langle \bm{h}_s^i, \bm{h}_p\rangle/\tau)},
\label{loss_sc}
\end{eqnarray}
% \end{small}
where $|\bm{H}_s| = K+\beta_s$ means the size of $\bm{H}_s$. Minimizing $\mathcal{L}_{\rm SC}$ encourages maximizing the distances between samples from different categories and clustering them from the same category closer together. Meanwhile, to bridge the distribution gap  between the prototypes of base categories and the features of the novel category, we employ the mean squared error (MSE) operation to measure the average prototypes and the novel features, and the loss is designed as:
\begin{eqnarray}
\mathcal{L}_{\rm MSE} = \frac{1}{N}\sum_{s=1}^{N}||\frac{1}{K}\sum_{i=1}^K\bm{f}_s^i - \frac{1}{\beta_s}\sum_{j=1}^{\beta_s}\bar{\bm{p}}_s^j||. 
\label{loss_mse}
\end{eqnarray}

Finally, we design a $N$-way classifier $\bm{\Gamma}$ to learn the prediction distribution from given novel features and the prototypes. In this work, the classifier $\bm{\Gamma}$ is a simple network, \emph{e.g.}, as simple as one fully connected layer. We use the cross-entropy loss to train the classifier with the hard labels: 
\begin{eqnarray}
\mathcal{L}_{\rm CE} =\frac{1}{N}\frac{1}{|\bm{H}_s|}\sum_{s=1}^{N}\sum_{i=1}^{|\bm{H}_s|}{\rm CrossEntropy}(\bm{h}_s^i, \bm{l}_s).
\end{eqnarray}

The total loss for training is defined as:
\begin{eqnarray}
\mathcal{L} = \mathcal{L}_{\rm CE} + \mu_1\mathcal{L}_{\rm SC} + \mu_2\mathcal{L}_{\rm MSE},
\label{eq:allloss}
\end{eqnarray}
where $\mu_1$ and $\mu_2$ are two weighting factors.

During the inference, we use the trained classifier to directly predict the category for each feature in the query set.

\section{Experiments}
 In this section, we present the experimental evaluation of our AFR. We begin by introducing the experimental settings. Next, we perform ablation studies to analyze the contributions of different components in our approach. Finally, we compare the performance of our approach with other state-of-the-art (SOTA) methods. Our experiments aim to address the following research questions (\textbf{RQ}):
 
\noindent\textbf{RQ1:} Given an novel category, how many related categories ($\mathcal{C}_{\beta_s}$) should be selected from the base categories?

\noindent\textbf{RQ2:} What are the effects of the instance attention and channel attention? 

\noindent\textbf{RQ3:} How do the contrastive learning and the feature space closing operations influence the classifier?

 \noindent\textbf{RQ4:} How does AFR perform compared to the state-of-the-art FSL methods?

\subsection{Experimental Settings}
\noindent\textbf{Datasets.} We evaluate our method on three benchmark datasets, \textit{i.e.}, Mini-ImageNet \cite{vinyals2016matching}, Tiered-ImageNet \cite{ren2018meta}, and Meta-Datase \cite{triantafillou2019meta}.
Specifically, Mini-ImageNet consists of 100 categories and each category has 600 images. It is divided into three parts: 64 base categories for training, 16 novel categories for validation, and the remaining 20 categories for testing. Similar to Mini-ImageNet, Tiered-ImageNet consists of 779165 images from 608 categories, where 351 base categories are used for training, 97 novel categories are used for validation, and the remaining 160 novel categories are used for testing. Meta-Dataset is a significantly larger-scale dataset that comprises multiple datasets with diverse data distributions, and we follow the usage described in \cite{xu2022alleviating}. Specifically, 
feature extractor training is conducted using the base categories of Mini-ImageNet, and the other 8 image datasets are utilized for testing process,  including Omniglot \cite{lake2015human}, CUB-200-2011 \cite{wah2011caltech}, Describable Textures \cite{CimpoiMKMV14},  Quick Draw \cite{fernandez2019quick}, Fungi \cite{SulcPMJH20}, VGG Flower \cite{NilsbackZ08}, Traffic Signs \cite{HoubenSSSI13}. 
\begin{figure}[t]
\centering
\includegraphics[width=0.48\textwidth]{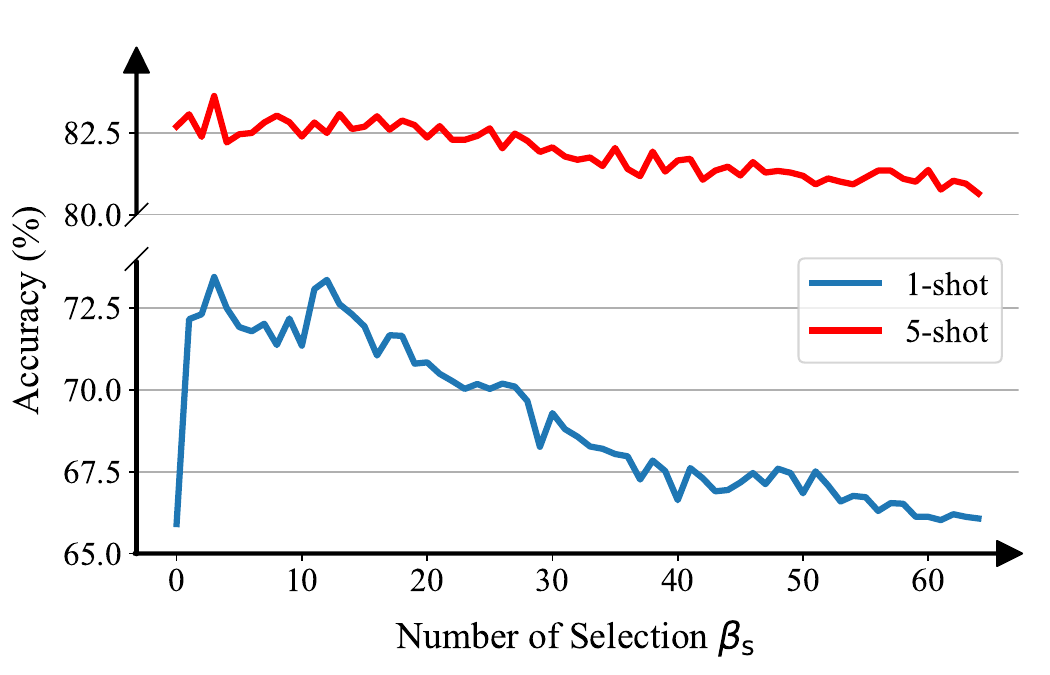}
\caption{The accuracy (\%) of the classifiers trained with the different numbers of selected base categories.}
\label{fig:select}
\end{figure}

\begin{table}[t]
\centering
\begin{tabular}{cc|cc}
\hline
{\textit{Ins.Att.}} & {\textit{Chanl.Att}}               & $K=1$                 & $K=5$         \\
\hline\hline
\XSolidBrush        & \XSolidBrush       & 64.02 $\pm$ 0.70\%           & 82.32 $\pm$ 0.41\%         \\
\XSolidBrush        & \CheckmarkBold     & 68.74 $\pm$ 0.61\%          & 82.56 $\pm$ 0.46\%           \\ 
\CheckmarkBold      & \XSolidBrush       & 68.03 $\pm$ 0.58\%           & 83.04 $\pm$ 0.43\%          \\ 
\CheckmarkBold      & \CheckmarkBold     & \textbf{70.68} $\pm$ 0.61\%  & \textbf{83.36} $\pm$ 0.45\%         \\ 
\hline
\end{tabular}
\caption{The accuracy (\%) of the classifiers with different attentions, where \textit{Ins.Att.} and \textit{Chanl.Att} is the instance attention and channel attention, respectively.} 
\label{tab:module_att}
\end{table}
\noindent\textbf{Evaluation.} In our evaluation, we conduct several $N$-way-$K$-shot classification tasks. In each task, $N$ novel categories are randomly sampled at first, then $K$ samples in each of the $N$ categories are sampled for training, and finally, 15 samples (different from the previous $K$ samples) in each of the $N$ categories are sampled for testing. To ensure reliable results, we sample 600 such tasks and report mean accuracies and variances on all tasks. In our experiments, $N=5$. Notably, we adhere to the evaluation setting for meta-dataset as described in \cite{xu2022alleviating}, where the novel categories are randomly sampled from the alternate image datasets, excluding the base categories present in the Mini-ImageNet.

\noindent\textbf{Implementation Details.} We utilize the features extracted from the pre-trained model and then apply our AFR to obtain both original and regularized features for training the classifier $\bm{\Gamma}$. These features are used to train the classifier $\bm{\Gamma}$ using the loss function $\mathcal{L}$ defined in Eq.~\eqref{eq:allloss} for a total of 1000 epochs. We employ Adam optimization \cite{kingma2015adam} with a learning rate of 0.001 and a weight decay of 0.0001 during the training process.

\subsection{Ablation Study}
In the ablation study, we use 64 base categories and 16 novel categories (validation set) of Mini-ImageNet with the available ResNet-12 \cite{chen2021meta} to evaluate the effectiveness of the different components of attentive feature regularization (AFR). Meanwhile, we use the pre-trained word2vec \cite{li2019large} to represent the labels with vectors. All experiments in the ablation study are conducted on 5-way-$K$-shot settings, where $K=1$ or $K=5$. We first evaluate the category selection and then introduce the experiments of different attention calculations and training strategies.

\subsubsection{The influences of semantic selection (\textbf{RQ1})}
The semantic selection is designed for feature regularization, thus we train the classifier $\Gamma$ with only instance attention and $\mathcal{L}_{\rm CE}$ loss to validate its effects. In this ablation study, we conduct experiments with different $\beta_s$ on $K=1$ and $K=5$, where $\beta_s$ ranges from 1 to 64 (base categories of Mini-ImageNet). The results are shown in Figure~\ref{fig:select}. For comparison, we also plot the results without any operation at $0^{\rm th}$ position (``Baseline''). First, both the introduced semantic selection and instance attention can improve the performance of the classifier. Moreover, the accuracy of using instance attention is better than that of utilizing the whole base categories ($64^{\rm th}$ position). 
Second, with the increase of category selection, the performances of the classifier increase first and then decrease. It's because introducing too many categories also bring more noise, which makes it hard to train the classifier. Therefore, we set $\beta_s=3$ in our remaining experiments.

\begin{table}[t]
\centering
\begin{tabular}{cc|cc}
\hline
{$\mathcal{L}_{\rm SC}$} & {$\mathcal{L}_{\rm MSE}$}    &    $ K=1$     &     $K=5$    \\
\hline\hline
\XSolidBrush                            & \XSolidBrush        & 70.68 $\pm$ 0.61\%           & 83.36 $\pm$ 0.45\%         \\
\XSolidBrush                            & \CheckmarkBold      & 70.93 $\pm$ 0.66\%           & 83.76 $\pm$ 0.43\%         \\
\CheckmarkBold                          & \XSolidBrush        & 71.18 $\pm$ 0.63\%           & 83.70 $\pm$ 0.45\%         \\
\CheckmarkBold                          & \CheckmarkBold      & \textbf{72.35} $\pm$ 0.63\%  & \textbf{84.11} $\pm$ 0.43\%         \\ 
\hline
\end{tabular}
\caption{The accuracy (\%) of the classifiers with different training strategies of loss functions.}
\label{tab:aba_loss}
\end{table}

\begin{table}[t]
\centering
\begin{tabular}{c|cc}
\hline
Regularization    &                     $K=1$               &    $K=5$         \\
\hline\hline
Baseline   & 64.02 $\pm$ 0.70\%           & 82.58 $\pm$ 0.45\%                 \\
CutMix$^\dag$    & 64.83 $\pm$ 0.72\%           & 80.89 $\pm$ 0.51\%           \\
Mixup$^\dag$     & 64.93 $\pm$ 0.69\%           & 81.55 $\pm$ 0.47\%           \\
CutOut$^\dag$    & 64.85 $\pm$ 0.68\%           & 81.53 $\pm$ 0.48\%           \\ 
\hline\hline
\textbf{AFR}$($only  $\bar{\bm{P}}_{S})$ &  67.58 $\pm$ 0.69\%           & 82.96 $\pm$ 0.46\% \\
\textbf{AFR}$(\bm{f}_{s} + \bar{\bm{P}}_{S})$  & \textbf{72.35} $\pm$ 0.63\%  & \textbf{83.74} $\pm$ 0.43\%         \\ 
\hline
\end{tabular}
\caption{The accuracy (\%) of the classifiers trained with different regularization strategies. $^\dag$ is our implementation.}
\label{other_reg}
\end{table}

\subsubsection{The effects of different attentions (\textbf{RQ2})}
To evaluate the effectiveness of different attentions, we train four classifiers with or without attention operations, using only the $\mathcal{L}_{\rm CE}$ loss. The performance of each classifier was evaluated on $K=1$ and $K=5$, and the results are shown in Table~\ref{tab:module_att}. The results indicate that both instance and channel attention improve the classifier performance for the query samples. Compared to the classifier without employing any attention, the introduced instance attention and channel attention achieve nearly 6\% accuracy improvements on the $K=1$ experiment, respectively. More importantly, combining these attentions provides the best performance (the last row of Table~\ref{tab:module_att}), with over 7.5\% improvement, which validates the effectiveness of our attention calculations.

\subsubsection{The effectiveness of different losses (\textbf{RQ3})} 
In this ablation study, we train four classifiers with different loss functions, where the instance attention and the channel attention are applied in all cases. To balance the optimization process of these losses, we set $\mu_1=5$ and $\mu_2=20$ experientially and following \cite{li2022selective}. The performances of four classifiers on $K=1$ and $K=5$ are shown in Table~\ref{tab:aba_loss}, which show that both $\mathcal{L}_{\rm SC}$ and $\mathcal{L}_{\rm MSE}$ contribute to the training procedure of the classifier. Moreover, combining these two losses further improves classification performance. 

We also verify the effects of different regularizations in Table~\ref{other_reg}. The common regularizations, \textit{i.e.} CutMix, Mixup, and CutOut, achieve slight improvement over the baseline in the 1-shot task but are harmful to accuracy in the 5-shot task.  
The classifier trained with the regularized features obtains over 3\% improvements (\textbf{AFR}$($only $\bar{\bm{P}}_{S})$). Moreover, our AFR $(\bm{f}_{s} + \bar{\bm{P}}_{S})$ can further improve the performances.

\begin{table*}[!t]
\centering
\begin{tabular}{l|c|cc|cc}
\hline
\multicolumn{1}{c|}{\multirow{2}*{\textbf{Method}}} & \multirow{2}*{\textbf{Backbone}} & \multicolumn{2}{c|}{\textbf{Mini-ImageNet}} & \multicolumn{2}{c}{\textbf{Tiered-ImageNet}}\\
& & $K=1$ & $K=5$ & $K=1$ & $K=5$   \\ 
\hline\hline
MatchingNets (NeurIPS16) & ResNet-12 & 63.08 $\pm$ 0.80\% & 75.99 $\pm$ 0.60\% & 68.50 $\pm$ 0.92\% & 80.60 $\pm$ 0.71\% \\
ProtoNets (NeurIPS17)  & ResNet-12 & 60.37 $\pm$ 0.83\% & 78.02 $\pm$ 0.57\% & 65.65 $\pm$ 0.92\% & 83.40 $\pm$ 0.65\% \\
MixtFSL (ICCV21) & ResNet-12 & 63.98 $\pm$ 0.79\% & 82.04 $\pm$ 0.49\% & 70.97 $\pm$ 1.03\% & 86.16 $\pm$ 0.67\% \\
RENet (ICCV21) & ResNet-12 & 67.60 $\pm$ 0.44\% & 82.58 $\pm$ 0.30\% & 71.61 $\pm$ 0.51\% & 85.28 $\pm$ 0.35\% \\
DeepBDC (CVPR22) & ResNet-12 & 67.34 $\pm$ 0.43\% & 84.46 $\pm$ 0.28\% & 72.34 $\pm$ 0.49\% & 87.31 $\pm$ 0.32\% \\
FeLMi (NeurIPS22) & ResNet-12 & 67.47 $\pm$ 0.78\% & 86.08 $\pm$ 0.44\% & 71.63 $\pm$ 0.89\% & 87.01 $\pm$ 0.55\% \\
tSF(ECCV22) & ResNet-12 & 69.74 $\pm$ 0.47\% & 83.91 $\pm$ 0.30\% & 71.89 $\pm$ 0.50\% & 85.49 $\pm$ 0.35\% \\
\hline
FEAT (CVPR20) & ResNet-12 & 66.78 $\pm$ 0.20\% & 82.05 $\pm$ 0.14\% & 70.80 $\pm$ 0.23\% & 84.79 $\pm$ 0.16\% \\
FEAT + \textbf{AFR} & ResNet-12 &   \textbf{72.57} $\pm$ 0.62\% &   \textbf{85.06} $\pm$ 0.42\%  &  \textbf{71.55} $\pm$ 0.74\% &  \textbf{87.64} $\pm$ 0.46\% \\
\hline
Meta-Baseline (ICCV21) & ResNet-12 & 63.17 $\pm$ 0.23\% & 79.26 $\pm$ 0.17\% & 68.62 $\pm$ 0.27\% & 83.74 $\pm$ 0.18\% \\
Meta-Baseline + \textbf{AFR}  & ResNet-12 &  \textbf{69.78} $\pm$  0.61\% &  \textbf{84.51} $\pm$ 0.41\%  &  \textbf{69.66} $\pm$ 0.70\% &  \textbf{86.29} $\pm$ 0.48\% \\
\hline
FRN (CVPR21) & ResNet-12 & 66.45 $\pm$ 0.19\% & 82.83 $\pm$ 0.13\% & 71.16 $\pm$ 0.22\% & 86.01 $\pm$ 0.15\% \\
FRN + \textbf{AFR}   & ResNet-12 &  \textbf{71.66} $\pm$ 0.56\% &  \textbf{84.75} $\pm$ 0.46\% &  \textbf{71.54} $\pm$ 0.71\% &  \textbf{87.35} $\pm$ 0.47\%  \\
\hline
BML (ICCV21) & ResNet-12 & 67.04 $\pm$ 0.63\% & 83.63 $\pm$ 0.29\% & 68.99 $\pm$ 0.50\% & 85.49 $\pm$ 0.34\% \\
BML + \textbf{AFR} & ResNet-12 &   \textbf{73.84} $\pm$  0.60\% &  \textbf{86.63} $\pm$ 0.41\% &  \textbf{73.41} $\pm$ 0.74\% & \textbf{87.44} $\pm$ 0.48\%  \\
\hline
% DeepEMD (CVPR20)  & ResNet-12 & 65.91 $\pm$ 0.82\% & 82.41 $\pm$ 0.56\% & 71.16 $\pm$ 0.87\% & 86.03 $\pm$ 0.58\% \\
% DeepEMD + \textbf{AFR} & ResNet-12 &   \textbf{72.42} $\pm$ 0.57\% &   \textbf{84.02} $\pm$ 0.44\%  &  \textbf{74.22} $\pm$ 0.69\% &  \textbf{88.66} $\pm$ 0.46\% \\
% \hline
Label-Halluc (AAAI22)  & ResNet-12 & 68.28 $\pm$ 0.77\% & 86.54 $\pm$ 0.46\% & 73.34 $\pm$ 1.25\% & 87.68 $\pm$ 0.83\% \\
Label-Halluc + \textbf{AFR} & ResNet-12 &   \textbf{74.57} $\pm$ 0.58\% &   \textbf{87.30} $\pm$ 0.37\%  &  \textbf{73.66} $\pm$ 0.66\% &  \textbf{89.15} $\pm$ 0.40\% \\
\hline
SEGA (WACV22)  & ResNet-12 & 69.04 $\pm$ 0.26\% & 79.03 $\pm$ 0.18\% & 72.18 $\pm$ 0.30\% & 84.28 $\pm$ 0.21\% \\
SEGA + \textbf{AFR} & ResNet-12 &   \textbf{71.14} $\pm$ 0.60\% &   \textbf{84.26} $\pm$ 0.42\%  &  \textbf{72.87} $\pm$ 0.45\% &  \textbf{85.26} $\pm$ 0.54\% \\
\hline\hline
IFSL (NeurIPS20) & WRN-28-10 & 64.12 $\pm$ 0.44\% & 80.97 $\pm$ 0.31\% & 69.96 $\pm$ 0.46\% & 86.19 $\pm$ 0.34\% \\
tSF (ECCV22) & WRN-28-10 & 70.23 $\pm$ 0.46\% & 84.55 $\pm$ 0.29\% & 74.87 $\pm$ 0.49\% & 88.05 $\pm$ 0.32\% \\
RankDNN (AAAI23)& WRN-28-10 & 66.67 $\pm$ 0.15\% & 84.79 $\pm$ 0.11\% & 74.00 $\pm$ 0.15\% & 88.80 $\pm$ 0.25\% \\
\hline
FEAT (CVPR20) & WRN-28-10 & 65.10 $\pm$ 0.20\% & 81.11 $\pm$ 0.14\% & 70.41 $\pm$ 0.23\% & 84.38 $\pm$ 0.16\% \\
FEAT + \textbf{AFR} & WRN-28-10 &   \textbf{71.76} $\pm$ 0.59\% &   \textbf{84.60} $\pm$ 0.42\%  &   \textbf{71.74} $\pm$ 0.74\% &  \textbf{86.33} $\pm$ 0.53\% \\
\hline
LRDC (ICLR21) & WRN-28-10 & 68.57 $\pm$ 0.55\% & 82.88 $\pm$ 0.42\% & 74.38$^\dag$ $\pm$ 0.93\% & 88.12$^\dag$ $\pm$ 0.59\% \\
LRDC + \textbf{AFR} & WRN-28-10 &   \textbf{72.98} $\pm$ 0.62\% &   \textbf{86.91} $\pm$ 0.40\%  &   \textbf{75.26} $\pm$ 0.67\% &  \textbf{89.59} $\pm$ 0.46\% \\
\hline
\end{tabular}
\caption{The accuracies (\%) by different methods on the novel categories from Mini-ImageNet \cite{vinyals2016matching} and Tiered-ImageNet \cite{ren2018meta}. $^\dag$ denotes our implementation.} 
\label{tab:all_imagenet}
\end{table*}

\begin{table*}[!t]
\centering
\begin{tabular}{l|cccccccc}
\hline
\multicolumn{1}{c|}{\multirow{2}{*}{\textbf{Method}}} & \multicolumn{8}{c}{\textbf{Testing Data Set}} \\ 
\cline{2-9}             & \multicolumn{1}{c}{Mini-Test} & \multicolumn{1}{c}{CUB} & \multicolumn{1}{c}{Fungi}   
                        &\multicolumn{1}{c}{Omini}     & \multicolumn{1}{c}{Sign} & \multicolumn{1}{c}{QDraw} 
                        & \multicolumn{1}{c}{Flower} & DTD \\ 
\hline\hline
SimpleShot (arXiv2019) & 67.18\% & 49.68\%   &   43.79\%   &  78.19\%  &  54.04\%  & 54.50\% &  71.68\%   &  51.19\%  \\
ZN (ICCV2021) & 67.05\% & 48.15\%    &   43.24\%  &  78.80\%  &  53.92\%  & 52.86\% &  72.01\%   &  52.20\% \\
TCPR (NeurIPS 2022) & 69.52\% & 53.83\%    &   46.28\%  &  80.88\%  &  56.65\%  & 57.31\% &  75.37\%   &  54.38\% \\
\hline\hline
\makecell[c]{\textbf{AFR}} & \textbf{72.98}\% & \textbf{54.45}\%   &   \textbf{47.93}\%  &  \textbf{81.84}\% &  \textbf{60.12}\%    &  \textbf{58.20}\%   &   \textbf{76.11}\% & \textbf{57.47}\% \\ 
\hline
\end{tabular}
\caption{The accuracies (\%) by different methods on Meta-Dataset \cite{triantafillou2019meta} with $K=1$. Sign and DTD denote Traffic Signs and Describable Textures dataset, respectively.} 
\label{tab:meata-dataset}
\end{table*}

\subsection{Comparisons with Other Methods (\textbf{RQ4})}
We compare the performance of our method with the latest on the \textbf{Mini-ImageNet} and \textbf{Tiered-ImageNet} datasets. Table~\ref{tab:all_imagenet} shows the results which contain MatchingNets \cite{lee2019meta},
ProtoNets \cite{snell2017prototypical},
%AM3 \cite{xing2019adaptive},
% MetaOptNet \cite{lee2019meta},
% Robust20-dist++ \cite{dvornik2019diversity},
% DMF \cite{xu2021learning},
MixtFSL \cite{afrasiyabi2021mixture},
RENet \cite{kang2021relational},
DeepBDC \cite{xie2022joint},
FeLMi \cite{roy2022felmi},
tSF \cite{lai2022tsf},
RankDNN \cite{guo2023rankdnn},
FRN \cite{wertheimer2021few},
BML \cite{zhou2021binocular},
FEAT \cite{ye2020few},
Label-Halluc \cite{jian2022label},
SEGA \cite{yang2022sega},
IFSL \cite{yue2020interventional},
and LDRC \cite{yang2021free}.
At the same time, we apply our approach to five recently proposed popular FSL methods, \textit{i.e.}, Meta-Baseline, FRN, BML, FEAT, Label-Halluc, SEGA, and LRDC. We clearly observe that our approach consistently improves the classification performance in all settings, which is agnostic to the method, datasets, and pre-trained backbones. For different features extracted with various methods on Mini-ImageNet, we perform remarkable 6.61\% accuracy improvements with the baseline (``Meta-Baseline + $\textbf{AFR}$'') and obtain the best accuracy 74.57\% with features from \cite{zhou2021binocular} (``Label-Halluc + $\textbf{AFR}$'') under $K=1$. Generally, our AFR outperforms the compared methods by about 2\% in accuracy for $K = 1$ and the improvements are generally greater in the 1-shot setting compared to the 5-shot setting. In the Tiered-ImageNet, we gain the  4.42\% improvement (``BML + $\textbf{AFR}$'') and achieve the best performance 89.59\% (``LRDC + $\textbf{AFR}$'') for $K=1$ and $K=5$, respectively.

To further demonstrate the effectiveness of our AFR, we conduct evaluations on the Meta-Dateset with $K=1$ setting. The results are summarized in Table~\ref{tab:meata-dataset}, including SimpleShot\cite{wang2019simpleshot},  ZN\cite{fei2021z}, and TCPR\cite{xu2022alleviating}. We can see that our AFR  exhibits strong adaptability to new data domains and achieves the best classification performance across several testing datasets. Notably, even compared to the transductive setting of TCPR, our approach gains more than 3\% improvements on Traffic Signs and Describale Textures datasets. 
% We conduct additional experiments to further evaluate the performance of our AFR method in the Technical Appendix. These experiments include comparisons with other methods on CIFAR-FS \cite{bertinetto2018meta} and CUB datasets \cite{wah2011caltech}. Additionally, we conduct t-SNE \cite{van2008visualizing} visualizations to demonstrate the effectiveness of our approach in a more intuitive manner.
\section{Conclusion}
In this paper, we have proposed attentive feature regularization named AFR to tackle the challenges in few-shot learning. Specifically, (1) The category selection based on semantic knowledge is employed to carefully constrain the features for regularization and helps avoid introducing unrelated noise into the training process.  (2) Two attention calculations are designed to improve the complementarity of the features across the different categories and improve the channel discriminability of the regularized features. 
The extensive experiments have demonstrated the effectiveness of our proposed method, particularly in the $1$-shot setting. 
% In our future work, we will focus on achieving a more robust feature regularization by incorporating additional techniques, such as GCN (Graph Convolutional Network) and GNN (Graph Neural Network), \textit{et al.}, to further enhance the performance of the classifier.

Note that the current usage of semantic relations is superficial. In our future work, we will focus on achieving a more robust feature regularization by incorporating additional techniques, such as GCN (Graph Convolutional Network) and GNN (Graph Neural Network), \textit{et al.}, to further enhance the performance of the classifier.
\section{Acknowledgements}
This work was supported by the National Natural Science Foundation of China (Grants No. 62202439).

\bibliography{main}

\begin{thebibliography}{60}
\providecommand{\natexlab}[1]{#1}

\bibitem[{Afrasiyabi, Lalonde, and Gagn{\'{e}}(2021)}]{afrasiyabi2021mixture}
Afrasiyabi, A.; Lalonde, J.; and Gagn{\'{e}}, C. 2021.
\newblock Mixture-based Feature Space Learning for Few-shot Image Classification.
\newblock In \emph{{ICCV}}, 9021--9031.

\bibitem[{Chen et~al.(2021)Chen, Liu, Xu, Darrell, and Wang}]{chen2021meta}
Chen, Y.; Liu, Z.; Xu, H.; Darrell, T.; and Wang, X. 2021.
\newblock Meta-Baseline: Exploring Simple Meta-Learning for Few-Shot Learning.
\newblock In \emph{{ICCV}}, 9042--9051.

\bibitem[{Chikontwe, Kim, and Park(2022)}]{chikontwe2022cad}
Chikontwe, P.; Kim, S.; and Park, S.~H. 2022.
\newblock {CAD:} Co-Adapting Discriminative Features for Improved Few-Shot Classification.
\newblock In \emph{{CVPR}}, 14534--14543.

\bibitem[{Chou et~al.(2020)Chou, Chang, Pan, Wei, and Juan}]{Remix}
Chou, H.; Chang, S.; Pan, J.; Wei, W.; and Juan, D. 2020.
\newblock Remix: Rebalanced Mixup.
\newblock In \emph{{ECCV} Workshops {(6)}}, volume 12540, 95--110.

\bibitem[{Cimpoi et~al.(2014)Cimpoi, Maji, Kokkinos, Mohamed, and Vedaldi}]{CimpoiMKMV14}
Cimpoi, M.; Maji, S.; Kokkinos, I.; Mohamed, S.; and Vedaldi, A. 2014.
\newblock Describing Textures in the Wild.
\newblock In \emph{CVPR}, 3606--3613.

\bibitem[{Deutsch et~al.(2017)Deutsch, Kolouri, Kim, Owechko, and Soatto}]{deutsch2017zero}
Deutsch, S.; Kolouri, S.; Kim, K.; Owechko, Y.; and Soatto, S. 2017.
\newblock Zero Shot Learning via Multi-scale Manifold Regularization.
\newblock In \emph{{CVPR}}, 5292--5299.

\bibitem[{Devries and Taylor(2017)}]{devries2017improved}
Devries, T.; and Taylor, G.~W. 2017.
\newblock Improved Regularization of Convolutional Neural Networks with Cutout.
\newblock \emph{CoRR}, abs/1708.04552.

\bibitem[{Fei et~al.(2021)Fei, Gao, Lu, and Xiang}]{fei2021z}
Fei, N.; Gao, Y.; Lu, Z.; and Xiang, T. 2021.
\newblock Z-score normalization, hubness, and few-shot learning.
\newblock In \emph{Proceedings of the IEEE/CVF International Conference on Computer Vision}, 142--151.

\bibitem[{Fernandez-Fernandez et~al.(2019)Fernandez-Fernandez, Victores, Estevez, and Balaguer}]{fernandez2019quick}
Fernandez-Fernandez, R.; Victores, J.~G.; Estevez, D.; and Balaguer, C. 2019.
\newblock Quick, stat!: A statistical analysis of the quick, draw! dataset.
\newblock \emph{arXiv preprint arXiv:1907.06417}.

\bibitem[{Guo, Mao, and Zhang(2019)}]{GuoMZ19}
Guo, H.; Mao, Y.; and Zhang, R. 2019.
\newblock MixUp as Locally Linear Out-of-Manifold Regularization.
\newblock In \emph{{AAAI}}, 3714--3722.

\bibitem[{Guo et~al.(2023)Guo, Haotong, Wei, Fu, Yu, Zhang, and Ge}]{guo2023rankdnn}
Guo, Q.; Haotong, G.; Wei, X.; Fu, Y.; Yu, Y.; Zhang, W.; and Ge, W. 2023.
\newblock RankDNN: Learning to Rank for Few-Shot Learning.
\newblock In \emph{Proceedings of the AAAI Conference on Artificial Intelligence}, volume~37, 728--736.

\bibitem[{Hariharan and Girshick(2017)}]{hariharan2017low}
Hariharan, B.; and Girshick, R.~B. 2017.
\newblock Low-Shot Visual Recognition by Shrinking and Hallucinating Features.
\newblock In \emph{{ICCV}}, 3037--3046.

\bibitem[{Hou et~al.(2019)Hou, Chang, Ma, Shan, and Chen}]{hou2019cross}
Hou, R.; Chang, H.; Ma, B.; Shan, S.; and Chen, X. 2019.
\newblock Cross Attention Network for Few-shot Classification.
\newblock In \emph{NeurIPS}, 4005--4016.

\bibitem[{Hou, Liu, and Wang(2017)}]{HouLW17}
Hou, S.; Liu, X.; and Wang, Z. 2017.
\newblock DualNet: Learn Complementary Features for Image Recognition.
\newblock In \emph{{ICCV}}, 502--510.

\bibitem[{Houben et~al.(2013)Houben, Stallkamp, Salmen, Schlipsing, and Igel}]{HoubenSSSI13}
Houben, S.; Stallkamp, J.; Salmen, J.; Schlipsing, M.; and Igel, C. 2013.
\newblock Detection of traffic signs in real-world images: The German traffic sign detection benchmark.
\newblock In \emph{{IJCNN}}, 1--8.

\bibitem[{Hu, Shen, and Sun(2018)}]{hu2018squeeze}
Hu, J.; Shen, L.; and Sun, G. 2018.
\newblock Squeeze-and-Excitation Networks.
\newblock In \emph{{CVPR}}, 7132--7141.

\bibitem[{Jian and Torresani(2022)}]{jian2022label}
Jian, Y.; and Torresani, L. 2022.
\newblock Label hallucination for few-shot classification.
\newblock In \emph{Proceedings of the AAAI Conference on Artificial Intelligence}, volume~36, 7005--7014.

\bibitem[{Kang et~al.(2021)Kang, Kwon, Min, and Cho}]{kang2021relational}
Kang, D.; Kwon, H.; Min, J.; and Cho, M. 2021.
\newblock Relational Embedding for Few-Shot Classification.
\newblock In \emph{{ICCV}}, 8802--8813.

\bibitem[{Khosla et~al.(2020)Khosla, Teterwak, Wang, Sarna, Tian, Isola, Maschinot, Liu, and Krishnan}]{khosla2020supervised}
Khosla, P.; Teterwak, P.; Wang, C.; Sarna, A.; Tian, Y.; Isola, P.; Maschinot, A.; Liu, C.; and Krishnan, D. 2020.
\newblock Supervised Contrastive Learning.
\newblock In \emph{NeurIPS}.

\bibitem[{Kingma and Ba(2015)}]{kingma2015adam}
Kingma, D.~P.; and Ba, J. 2015.
\newblock Adam: {A} Method for Stochastic Optimization.
\newblock In \emph{{ICLR}}.

\bibitem[{Lai et~al.(2022)Lai, Yang, Liu, Zeng, Huang, Wu, Liu, Gao, and Wang}]{lai2022tsf}
Lai, J.; Yang, S.; Liu, W.; Zeng, Y.; Huang, Z.; Wu, W.; Liu, J.; Gao, B.; and Wang, C. 2022.
\newblock tSF: Transformer-Based Semantic Filter for Few-Shot Learning.
\newblock In \emph{ECCV}.

\bibitem[{Lake, Salakhutdinov, and Tenenbaum(2015)}]{lake2015human}
Lake, B.~M.; Salakhutdinov, R.; and Tenenbaum, J.~B. 2015.
\newblock Human-level concept learning through probabilistic program induction.
\newblock \emph{Science}, 1332--1338.

\bibitem[{Lee et~al.(2019)Lee, Maji, Ravichandran, and Soatto}]{lee2019meta}
Lee, K.; Maji, S.; Ravichandran, A.; and Soatto, S. 2019.
\newblock Meta-Learning With Differentiable Convex Optimization.
\newblock In \emph{{CVPR}}, 10657--10665.

\bibitem[{Li et~al.(2020)Li, Huang, Lan, Feng, Li, and Wang}]{Limargin}
Li, A.; Huang, W.; Lan, X.; Feng, J.; Li, Z.; and Wang, L. 2020.
\newblock Boosting Few-Shot Learning With Adaptive Margin Loss.
\newblock In \emph{{CVPR}}, 12573--12581.

\bibitem[{Li et~al.(2019)Li, Luo, Lu, Xiang, and Wang}]{li2019large}
Li, A.; Luo, T.; Lu, Z.; Xiang, T.; and Wang, L. 2019.
\newblock Large-Scale Few-Shot Learning: Knowledge Transfer With Class Hierarchy.
\newblock In \emph{{CVPR}}, 7212--7220.

\bibitem[{Li et~al.(2022)Li, Xia, Ge, and Liu}]{li2022selective}
Li, S.; Xia, X.; Ge, S.; and Liu, T. 2022.
\newblock Selective-Supervised Contrastive Learning with Noisy Labels.
\newblock In \emph{{CVPR}}, 316--325.

\bibitem[{Liu et~al.(2021)Liu, Fu, Xu, Yang, Li, Wang, and Zhang}]{liu2021learning}
Liu, C.; Fu, Y.; Xu, C.; Yang, S.; Li, J.; Wang, C.; and Zhang, L. 2021.
\newblock Learning a Few-shot Embedding Model with Contrastive Learning.
\newblock In \emph{{AAAI}}, 8635--8643.

\bibitem[{Liu et~al.(2019)Liu, Zhao, Zhang, Cheng, and Zhu}]{LiuZZCZ19}
Liu, N.; Zhao, Q.; Zhang, N.; Cheng, X.; and Zhu, J. 2019.
\newblock Pose-Guided Complementary Features Learning for Amur Tiger Re-Identification.
\newblock In \emph{{ICCV} Workshops}, 286--293.

\bibitem[{Lu et~al.(2023)Lu, Wang, Zhang, Hao, and He}]{Lumm}
Lu, J.; Wang, S.; Zhang, X.; Hao, Y.; and He, X. 2023.
\newblock Semantic-based Selection, Synthesis, and Supervision for Few-shot Learning.
\newblock In \emph{{ACM} Multimedia}, 3569--3578.

\bibitem[{Luo, Xu, and Xu(2022)}]{luo2022channel}
Luo, X.; Xu, J.; and Xu, Z. 2022.
\newblock Channel Importance Matters in Few-Shot Image Classification.
\newblock In \emph{{ICML}}, volume 162 of \emph{Proceedings of Machine Learning Research}, 14542--14559. {PMLR}.

\bibitem[{Nilsback and Zisserman(2008)}]{NilsbackZ08}
Nilsback, M.; and Zisserman, A. 2008.
\newblock Automated Flower Classification over a Large Number of Classes.
\newblock In \emph{{ICVGIP}}, 722--729.

\bibitem[{Peng et~al.(2019)Peng, Li, Zhang, Li, Qi, and Tang}]{peng2019few}
Peng, Z.; Li, Z.; Zhang, J.; Li, Y.; Qi, G.; and Tang, J. 2019.
\newblock Few-Shot Image Recognition With Knowledge Transfer.
\newblock In \emph{{ICCV}}, 441--449.

\bibitem[{Ren et~al.(2018)Ren, Triantafillou, Ravi, Snell, Swersky, Tenenbaum, Larochelle, and Zemel}]{ren2018meta}
Ren, M.; Triantafillou, E.; Ravi, S.; Snell, J.; Swersky, K.; Tenenbaum, J.~B.; Larochelle, H.; and Zemel, R.~S. 2018.
\newblock Meta-Learning for Semi-Supervised Few-Shot Classification.
\newblock In \emph{{ICLR}}.

\bibitem[{Rodr{\'{\i}}guez et~al.(2020)Rodr{\'{\i}}guez, Laradji, Drouin, and Lacoste}]{rodriguez2020embedding}
Rodr{\'{\i}}guez, P.; Laradji, I.~H.; Drouin, A.; and Lacoste, A. 2020.
\newblock Embedding Propagation: Smoother Manifold for Few-Shot Classification.
\newblock In \emph{ECCV}, 121--138.

\bibitem[{Roy et~al.(2022)Roy, Shah, Shah, Dhar, Cherian, and Chellappa}]{roy2022felmi}
Roy, A.; Shah, A.; Shah, K.; Dhar, P.; Cherian, A.; and Chellappa, R. 2022.
\newblock FeLMi: Few shot Learning with hard Mixup.
\newblock In \emph{NeurIPS}.

\bibitem[{Shi, Wu, and Wang(2023)}]{ShiWW23a}
Shi, C.; Wu, H.; and Wang, L. 2023.
\newblock A Feature Complementary Attention Network Based on Adaptive Knowledge Filtering for Hyperspectral Image Classification.
\newblock \emph{{IEEE} Trans. Geosci. Remote. Sens.}, 61: 1--19.

\bibitem[{Snell, Swersky, and Zemel(2017)}]{snell2017prototypical}
Snell, J.; Swersky, K.; and Zemel, R.~S. 2017.
\newblock Prototypical Networks for Few-shot Learning.
\newblock In \emph{{NeurIPS}}, 4077--4087.

\bibitem[{Sulc et~al.(2020)Sulc, Picek, Matas, Jeppesen, and Heilmann{-}Clausen}]{SulcPMJH20}
Sulc, M.; Picek, L.; Matas, J.; Jeppesen, T.~S.; and Heilmann{-}Clausen, J. 2020.
\newblock Fungi Recognition: {A} Practical Use Case.
\newblock In \emph{{WACV}}, 2305--2313.

\bibitem[{Triantafillou et~al.(2019)Triantafillou, Zhu, Dumoulin, Lamblin, Evci, Xu, Goroshin, Gelada, Swersky, Manzagol et~al.}]{triantafillou2019meta}
Triantafillou, E.; Zhu, T.; Dumoulin, V.; Lamblin, P.; Evci, U.; Xu, K.; Goroshin, R.; Gelada, C.; Swersky, K.; Manzagol, P.-A.; et~al. 2019.
\newblock Meta-dataset: A dataset of datasets for learning to learn from few examples.
\newblock \emph{arXiv preprint arXiv:1903.03096}.

\bibitem[{Vaswani et~al.(2017{\natexlab{a}})Vaswani, Shazeer, Parmar, Uszkoreit, Jones, Gomez, Kaiser, and Polosukhin}]{VaswaniSPUJGKP17}
Vaswani, A.; Shazeer, N.; Parmar, N.; Uszkoreit, J.; Jones, L.; Gomez, A.~N.; Kaiser, L.; and Polosukhin, I. 2017{\natexlab{a}}.
\newblock Attention is All you Need.
\newblock In \emph{{NeurIPS}}, 5998--6008.

\bibitem[{Vaswani et~al.(2017{\natexlab{b}})Vaswani, Shazeer, Parmar, Uszkoreit, Jones, Gomez, Kaiser, and Polosukhin}]{vaswani2017attention}
Vaswani, A.; Shazeer, N.; Parmar, N.; Uszkoreit, J.; Jones, L.; Gomez, A.~N.; Kaiser, L.; and Polosukhin, I. 2017{\natexlab{b}}.
\newblock Attention is All you Need.
\newblock In \emph{{NeurIPS}}, 5998--6008.

\bibitem[{Velazquez et~al.(2022)Velazquez, Rodr{\i}guez, Gonfaus, Roca, and Gonzalez}]{velazquez2022closer}
Velazquez, D.; Rodr{\i}guez, P.; Gonfaus, J.~M.; Roca, F.~X.; and Gonzalez, J. 2022.
\newblock A Closer Look at Embedding Propagation for Manifold Smoothing.
\newblock \emph{The Journal of Machine Learning Research}, 23: 1--27.

\bibitem[{Verma et~al.(2019)Verma, Lamb, Beckham, Najafi, Mitliagkas, Lopez{-}Paz, and Bengio}]{verma2019manifold}
Verma, V.; Lamb, A.; Beckham, C.; Najafi, A.; Mitliagkas, I.; Lopez{-}Paz, D.; and Bengio, Y. 2019.
\newblock Manifold Mixup: Better Representations by Interpolating Hidden States.
\newblock In \emph{{ICML}}, 6438--6447.

\bibitem[{Vinyals et~al.(2016)Vinyals, Blundell, Lillicrap, Kavukcuoglu, and Wierstra}]{vinyals2016matching}
Vinyals, O.; Blundell, C.; Lillicrap, T.; Kavukcuoglu, K.; and Wierstra, D. 2016.
\newblock Matching Networks for One Shot Learning.
\newblock In \emph{{NeurIPS}}, 3630--3638.

\bibitem[{Wah et~al.(2011)Wah, Branson, Welinder, Perona, and Belongie}]{wah2011caltech}
Wah, C.; Branson, S.; Welinder, P.; Perona, P.; and Belongie, S. 2011.
\newblock The caltech-ucsd birds-200-2011 dataset.

\bibitem[{Wang et~al.(2020)Wang, Yue, Liu, Tian, and Wang}]{wang2020large}
Wang, S.; Yue, J.; Liu, J.; Tian, Q.; and Wang, M. 2020.
\newblock Large-Scale Few-Shot Learning via Multi-modal Knowledge Discovery.
\newblock In \emph{{ECCV}}, 718--734.

\bibitem[{Wang et~al.(2022)Wang, Zhang, Hao, Wang, and He}]{wang2022multi}
Wang, S.; Zhang, X.; Hao, Y.; Wang, C.; and He, X. 2022.
\newblock Multi-directional Knowledge Transfer for Few-Shot Learning.
\newblock In \emph{{ACM} Multimedia}, 3993--4002.

\bibitem[{Wang et~al.(2019)Wang, Chao, Weinberger, and Van Der~Maaten}]{wang2019simpleshot}
Wang, Y.; Chao, W.-L.; Weinberger, K.~Q.; and Van Der~Maaten, L. 2019.
\newblock Simpleshot: Revisiting nearest-neighbor classification for few-shot learning.
\newblock \emph{arXiv preprint arXiv:1911.04623}.

\bibitem[{Wang et~al.(2018)Wang, Girshick, Hebert, and Hariharan}]{wang2018low}
Wang, Y.; Girshick, R.~B.; Hebert, M.; and Hariharan, B. 2018.
\newblock Low-Shot Learning From Imaginary Data.
\newblock In \emph{{CVPR}}, 7278--7286.

\bibitem[{Wertheimer et~al.(2021)Wertheimer, Tang, Hariharan, , and and}]{wertheimer2021few}
Wertheimer, D.; Tang, L.; Hariharan, B.; ; and and. 2021.
\newblock Few-Shot Classification With Feature Map Reconstruction Networks.
\newblock In \emph{{CVPR}}, 8012--8021.

\bibitem[{Xie et~al.(2022)Xie, Long, Lv, Wang, and Li}]{xie2022joint}
Xie, J.; Long, F.; Lv, J.; Wang, Q.; and Li, P. 2022.
\newblock Joint Distribution Matters: Deep Brownian Distance Covariance for Few-Shot Classification.
\newblock In \emph{{CVPR}}, 7962--7971.

\bibitem[{Xu et~al.(2022)Xu, Luo, Pan, Li, Pei, and Xu}]{xu2022alleviating}
Xu, J.; Luo, X.; Pan, X.; Li, Y.; Pei, W.; and Xu, Z. 2022.
\newblock Alleviating the sample selection bias in few-shot learning by removing projection to the centroid.
\newblock \emph{Advances in Neural Information Processing Systems}, 35: 21073--21086.

\bibitem[{Yang, Wang, and Chen(2022)}]{yang2022sega}
Yang, F.; Wang, R.; and Chen, X. 2022.
\newblock SEGA: Semantic guided attention on visual prototype for few-shot learning.
\newblock In \emph{Proceedings of the IEEE/CVF Winter Conference on Applications of Computer Vision}, 1056--1066.

\bibitem[{Yang, Liu, and Xu(2021)}]{yang2021free}
Yang, S.; Liu, L.; and Xu, M. 2021.
\newblock Free Lunch for Few-shot Learning: Distribution Calibration.
\newblock In \emph{{ICLR}}.

\bibitem[{Ye et~al.(2020)Ye, Hu, Zhan, and Sha}]{ye2020few}
Ye, H.; Hu, H.; Zhan, D.; and Sha, F. 2020.
\newblock Few-Shot Learning via Embedding Adaptation With Set-to-Set Functions.
\newblock In \emph{{CVPR}}, 8805--8814.

\bibitem[{Yue et~al.(2020)Yue, Zhang, Sun, and Hua}]{yue2020interventional}
Yue, Z.; Zhang, H.; Sun, Q.; and Hua, X. 2020.
\newblock Interventional Few-Shot Learning.
\newblock In \emph{NeurIPS}.

\bibitem[{Yun et~al.(2019)Yun, Han, Chun, Oh, Yoo, and Choe}]{yun2019cutmix}
Yun, S.; Han, D.; Chun, S.; Oh, S.~J.; Yoo, Y.; and Choe, J. 2019.
\newblock CutMix: Regularization Strategy to Train Strong Classifiers With Localizable Features.
\newblock In \emph{{ICCV}}, 6022--6031.

\bibitem[{Zhang et~al.(2018)Zhang, Ciss{\'{e}}, Dauphin, and Lopez{-}Paz}]{zhang2018mixup}
Zhang, H.; Ciss{\'{e}}, M.; Dauphin, Y.~N.; and Lopez{-}Paz, D. 2018.
\newblock mixup: Beyond Empirical Risk Minimization.
\newblock In \emph{{ICLR}}.

\bibitem[{Zhou et~al.(2021)Zhou, Qiu, Xie, Wu, and Zhang}]{zhou2021binocular}
Zhou, Z.; Qiu, X.; Xie, J.; Wu, J.; and Zhang, C. 2021.
\newblock Binocular Mutual Learning for Improving Few-shot Classification.
\newblock In \emph{{ICCV}}, 8382--8391.

\bibitem[{Zhu et~al.(2023)Zhu, Zhang, He, Zhou, Wang, Zhao, and Gao}]{zhu2023not}
Zhu, X.; Zhang, R.; He, B.; Zhou, A.; Wang, D.; Zhao, B.; and Gao, P. 2023.
\newblock Not all features matter: Enhancing few-shot clip with adaptive prior refinement.
\newblock \emph{arXiv preprint arXiv:2304.01195}.

\end{thebibliography}

\end{document}